\documentclass[preprint,12pt]{elsarticle}

\usepackage{amsmath}
\usepackage{amssymb}
\usepackage{graphicx}
\usepackage{booktabs}
\usepackage{tabularx}
\usepackage{multirow}
\usepackage{array}
\usepackage{adjustbox}
\usepackage{subcaption}
\usepackage{enumitem}
\usepackage{microtype}
\usepackage{placeins}
\usepackage{cuted}
\usepackage{xurl}
\usepackage[hidelinks]{hyperref}

\biboptions{sort&compress}
\setlist[itemize]{leftmargin=*,nosep}
\newcolumntype{Y}{>{\raggedright\arraybackslash}X}
\newcolumntype{C}[1]{>{\centering\arraybackslash}p{#1}}

\journal{Data in Brief}

\begin{document}

\begin{frontmatter}

\title{Multimodal examination answer data with expert-designed Outcome-Based Education rubrics for criterion-level assessment}

\author[1]{Jahangir Alam SM\corref{cor1}}
\ead{jahangir@baust.edu.bd}
\author[1]{Md Khalid Syfullah\fnref{equal}}
\author[1]{Saad Ahmed\fnref{equal}}
\author[1]{Munira Akter Mou\fnref{equal}}
\author[1]{A K Z Rasel Rahman\fnref{equal}}
\author[1]{A.K.M. Masudur Rahman}
\author[1]{Mohammed Sowket Ali}

\fntext[equal]{These authors contributed equally to this work.}
\cortext[cor1]{Corresponding author}
\affiliation[1]{organization={Department of Computer Science and Engineering (CSE), Bangladesh Army University of Science and Technology (BAUST)}, city={Saidpur}, country={Bangladesh}}

\begin{abstract}
This data article describes a multimodal collection of scanned examination answers paired with expert-designed Outcome-Based Education (OBE) grading metadata. The collection contains 485 answer submissions from 415 consenting students at four academic institutions. Eight faculty contributors supplied examination materials covering nine subjects and 12 distinct question templates. Each answer-level item links a scanned PDF to a randomized identifier, subject label, question, model answer, criterion definitions, performance-level descriptions, criterion marks, and a total mark. The 12 rubrics contain 47 criteria in total. The scans retain realistic academic content, including handwriting, printed text, equations, tables, code, figures, sketches, and diagrams. CamScanner, Adobe Scan, and conventional scanners contributed variation in illumination, contrast, orientation, compression, and resolution. Diverse handwriting, crossed-out work, revised calculations, and inserted corrections add further visual variability for robustness and generalization studies. Preparation involved heterogeneous-source consolidation, label and text standardization, score validation, identifier randomization, filename randomization, and JSON-to-PDF integrity checks. An answer-level audit confirmed 485 unique identifiers, 485 unique PDF filenames, agreement between each total mark and its criterion-mark sum, and scores within the applicable rubric maximum. The data can support rubric-aware automated evaluation, multimodal document understanding, criterion-level feedback, score prediction, and privacy-aware OBE assessment research. Access is restricted to research use and is available from the corresponding author upon reasonable request.

\end{abstract}

\begin{keyword}
Outcome-Based Education \sep automated exam evaluation \sep scanned answer scripts \sep grading rubrics \sep document understanding \sep vision-language models \sep criterion-level assessment
\end{keyword}

\end{frontmatter}

\vspace{0.4em}
\begin{strip}
\section*{Specifications Table}
\centering
\captionof{table}{Specifications of the examination-answer dataset.}
\label{tab:specifications}
\footnotesize
\renewcommand{\arraystretch}{0.92}
\begin{tabularx}{\textwidth}{@{}p{0.20\textwidth}Y@{}}
\toprule
\textbf{Subject} & \textbf{Description} \\
\midrule
Subject area & Artificial intelligence, education, document understanding, and automated assessment \\
Specific subject area & Rubric-aware evaluation of scanned examination answers in an Outcome-Based Education framework \\
Data type and format & Scanned PDFs paired with answer-level JSON metadata, questions, model answers, rubric criteria, criterion marks, and total marks \\
Data collection & 485 answers from 415 consenting students, supplied by eight faculty contributors from four institutions \\
Parameters & Nine subjects, 12 question templates, 47 rubric criteria, and question maxima from 5 to 16 marks \\
Data source location & Four academic institutions; identifying institutional details are withheld to reduce identification risk \\
Data accessibility & Data are hosted on Zenodo \href{https://doi.org/10.5281/zenodo.22058762}{(DOI: 10.5281/zenodo.22058762)}. Access is restricted to research use and granted upon reasonable request through the repository, subject to the privacy safeguards stated in this article. \\
Related research article & Not applicable \\
\bottomrule
\end{tabularx}
\par
\end{strip}

\section{Value of the Data}

\begin{itemize}
\item The collection links visual examination evidence with question context, model answers, expert-authored OBE criteria, performance descriptions, criterion marks, and total marks at the answer level.
\item The nine-subject coverage and mixed visual content support research on handwritten and printed document understanding, optical character recognition, vision-language modeling, and cross-subject generalization.
\item Variation in lighting, handwriting, corrections, page geometry, and capture pipelines spanning CamScanner, Adobe Scan, and conventional scanners supports robustness studies under realistic acquisition conditions.
\item Criterion-level labels permit development of systems that report which parts of an answer satisfy examiner expectations, extending automated grading beyond a single aggregate score.
\item The paired PDF-JSON structure supports controlled experiments in score prediction, rubric selection, evidence grounding, feedback generation, calibration, and auditability.
\item The preparation workflow provides a reproducible pattern for consolidating heterogeneous examiner files, validating scores, randomizing record identifiers, and preserving file-to-record mappings.
\item Education researchers, document-AI researchers, assessment specialists, and institutions developing OBE workflows can reuse the data under approved research-only access conditions.
\end{itemize}

\section{Background}

Outcome-Based Education organizes curriculum, instruction, and assessment around explicit capabilities that learners are expected to demonstrate \citep{spady1994,biggs1996}. Assessment within this framework requires a visible connection among a question, the intended learning evidence, the criteria used by an examiner, and the marks assigned to the response. An overall score records final attainment in compact form. It does not preserve which required components were demonstrated, omitted, or expressed with partial quality.

Analytic rubrics address this limitation by dividing a task into criteria and describing performance at multiple quality levels. Clear, focused, task-specific rubrics can strengthen the consistency and interpretability of performance assessment \citep{jonsson2007,brookhart2015}. A rubric-aware dataset needs more than images and total marks. It must retain criterion definitions, performance descriptions, maximum allocations, criterion-level marks, and the link to the relevant question and reference answer.

Automatic short-answer grading has traditionally emphasized typed responses and aggregate labels. Prior reviews describe a progression from rule-based systems to statistical and neural approaches, with substantial variation in tasks and evaluation settings \citep{burrows2015}. Real examination scripts introduce a broader document-understanding problem. Student work may contain handwriting, equations, tables, code, graphs, or diagrams. Page layout and visual evidence can influence interpretation. Modern document-AI models combine text, layout, and image information \citep{huang2022}, while optical-character-recognition-free architectures learn direct mappings from document images to structured outputs \citep{kim2022}. Such systems require data that preserve both page appearance and structured assessment context.

The present collection was assembled for this purpose. Its unit of organization is one complete answer submission. Each submission retains the scanned PDF and an answer-level JSON record. The record supplies the subject, question, model answer, total mark, and a list of rubrics with criterion text, gained marks, and performance-level marking rules. This structure supports transparent experiments in which predicted scores can be traced to criterion-specific evidence.

The data article focuses on the data, its organization, and its preparation. It does not report model comparisons or operational grading performance. The dataset is intended for controlled research and is not intended to determine official student grades.

\section{Data Description}

\subsection{Collection overview}

The collection \cite{alam2026obeexam} contains 485 answer submissions associated with 415 participating students from four institutions. The difference between the two counts reflects answer-level organization: a participant can contribute more than one answer submission. Eight faculty contributors supplied the scanned scripts and assessment metadata. The coverage includes Machine Learning, Digital Image Processing, Database Systems, Computer Networks, Data Mining, Algorithms, Fisheries, Object Oriented Programming, and E-commerce.

\begin{figure*}[!t]
\centering
\includegraphics[width=\textwidth]{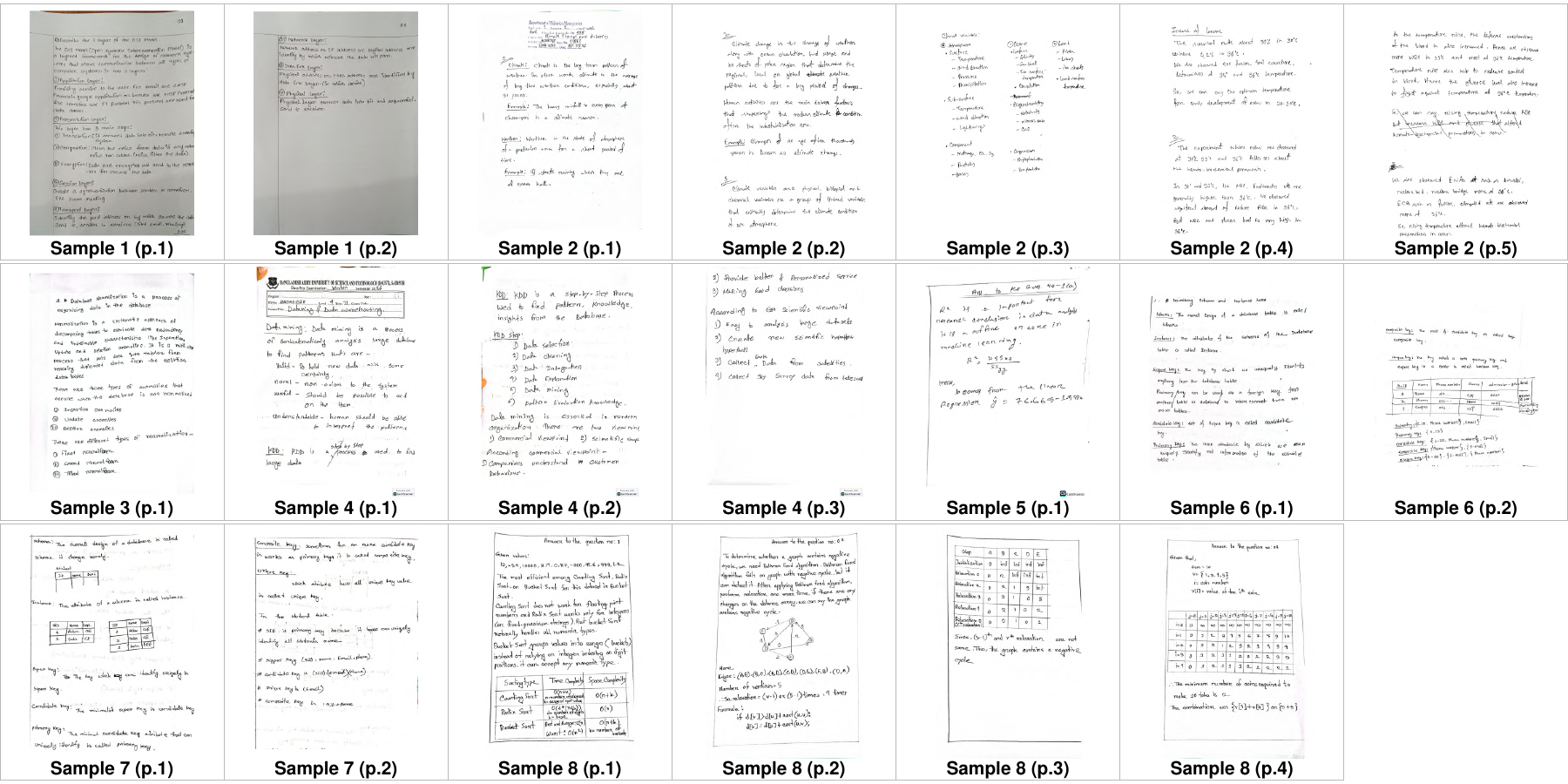}
\caption{Representative scanned answer pages from the dataset. The overview shows variation in handwriting, page count, equations, tables, diagrams, and institutional document formats.}
\label{fig:overview}
\end{figure*}

Each scan retains the original page appearance. The visual content includes plain text, mathematical notation, tables, figures, diagrams, code, and mixed layouts. Individual files vary in length. Some scripts contain one response, while others contain several questions or subquestions. Figure~\ref{fig:overview} presents representative page samples from the collection.

The scans were acquired with CamScanner, Adobe Scan, and conventional document scanners. Mobile capture and flatbed or sheet-fed scanning introduced different lighting conditions, brightness and contrast levels, shadows, page orientation, cropping, compression, and resolution. The scripts also preserve distinct handwriting styles, crossed-out text, overwritten symbols, corrected calculations, arrows, insertions, and other authentic revision marks. This combination broadens the visual domain represented by the dataset and supports evaluation of generalization across writing styles, document content, and acquisition conditions. External validation remains necessary before conclusions are extended beyond the participating institutions and subjects.

\subsection{Answer-level files and metadata}

The final organization uses paired visual and structured records. One randomized PDF filename identifies the scanned response, and one JSON object contains its metadata. Each object stores \texttt{answer\_id}, \texttt{pdf\_file}, \texttt{Subject}, \texttt{question}, \texttt{model\_answer}, and \texttt{total\_marks}. The \texttt{rubrics} field is a variable-length array whose length equals the number of criteria defined for the associated question. Each rubric element stores the criterion text, its \texttt{gained\_marks}, and the performance descriptions under \texttt{marking\_rules}. The four performance keys are \texttt{Excellent}, \texttt{Good}, \texttt{Average}, and \texttt{Poor}.

\begin{figure}[!t]
\centering
\includegraphics[width=0.7\columnwidth]{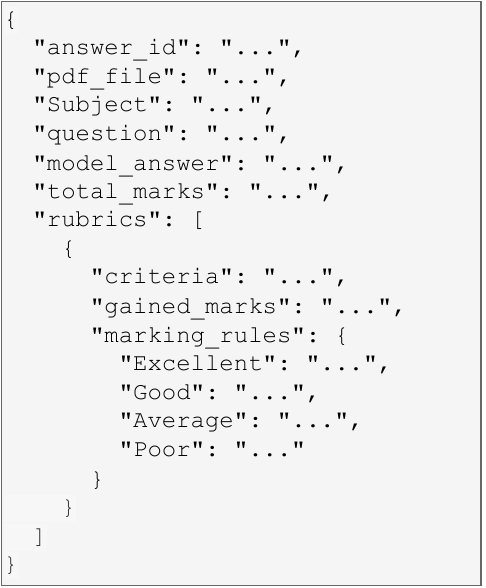}
\caption{Answer-level JSON representation linking a randomized scan with its question, reference material, rubric criteria, obtained marks, and performance-level rules.}
\label{fig:json}
\end{figure}

The JSON structure keeps the grading context next to the file reference, as illustrated in Figure~\ref{fig:json}. It supports answer-level loading without joining separate question, rubric, and mark tables. The underlying PDF remains unchanged as visual evidence. This arrangement is suitable for pipelines that render pages, extract text, encode document images, or combine both modalities.

\subsection{Subject distribution}

Table~\ref{tab:subjects} summarizes the subject distribution derived from the workbook and verified against the final JSON. Dataset shares are calculated from 485 answers. Mean marks are normalized by the maximum of each answer's question rubric prior to subject-level averaging. This normalization makes questions with 5, 6, 9, 10, or 16 available marks comparable within a subject.

\begin{table*}[!t]
\centering
\caption{Subject-level composition. Mean normalized mark is descriptive metadata rather than a model evaluation.}
\label{tab:subjects}
\footnotesize
\renewcommand{\arraystretch}{1.1}
\begin{tabularx}{\textwidth}{@{}Y C{1.3cm} C{1.35cm} C{1.4cm} C{1.7cm} p{4.1cm}@{}}
\toprule
\textbf{Subject} & \textbf{Answers} & \textbf{Questions} & \textbf{Share (\%)} & \textbf{Mean normalized mark (\%)} & \textbf{Visual answer content} \\
\midrule
Machine Learning & 88 & 2 & 18.14 & 77.90 & Primarily text \\
Digital Image Processing & 75 & 2 & 15.46 & 77.20 & Primarily text \\
Database Systems & 80 & 2 & 16.49 & 62.91 & Text, tables, equations \\
Computer Networks & 70 & 1 & 14.43 & 64.57 & Text, optional diagrams \\
Data Mining & 42 & 1 & 8.66 & 56.40 & Primarily text \\
Algorithms & 83 & 1 & 17.11 & 82.98 & Text, equations, diagrams \\
Fisheries & 23 & 1 & 4.74 & 91.49 & Primarily text \\
Object Oriented Programming & 19 & 1 & 3.92 & 53.45 & Text and code \\
E-commerce & 5 & 1 & 1.03 & 48.89 & Primarily text \\
\midrule
\textbf{Total} & \textbf{485} & \textbf{12} & \textbf{100.00} & & \\
\bottomrule
\end{tabularx}
\end{table*}

\subsection{Question and rubric structure}

The 12 question templates contain 47 criteria. Nine questions use four criteria, two use three criteria, and one uses five criteria. Question maxima range from 5 to 16 marks. Table~\ref{tab:questions} reports the answer type, number of answers, criterion count, rubric maximum, mean total mark, and observed mark range for each question. The verified answer-level JSON supplies the descriptive values, while the workbook supplies the answer-type categories.

\begin{table*}[!t]
\centering
\caption{Question-level grading structure, answer type, and mark coverage.}
\label{tab:questions}
\scriptsize
\renewcommand{\arraystretch}{1.05}
\begin{tabularx}{\textwidth}{@{}C{0.45cm} >{\raggedright\arraybackslash}p{2.35cm} Y >{\raggedright\arraybackslash}p{1.90cm} C{0.90cm} C{0.90cm} C{0.96cm} C{0.86cm}@{}}
\toprule
\textbf{Q} & \textbf{Subject and focus} & \textbf{Question summary} & \textbf{Answer type} & \textbf{Answers} & \textbf{Criteria} & \textbf{Mean mark} & \textbf{Range} \\
\midrule
1 & Machine Learning, $R^2$ & Importance, mathematical interpretation, limitations, and model comparison & Plain text & 25 & 4 & 2.82/5 & 0 to 5 \\
2 & Machine Learning, paradigms & Supervised, unsupervised, and reinforcement learning & Plain text & 63 & 4 & 8.64/10 & 0 to 10 \\
3 & Digital Image Processing, quantization & Purpose of quantization and the Max-Lloyd algorithm & Plain text & 40 & 4 & 8.53/10 & 2 to 10 \\
4 & Digital Image Processing, histograms & Histogram equalization and histogram matching & Plain text & 35 & 4 & 6.80/10 & 4 to 10 \\
5 & Database Systems, normalization & Redundancy, update anomalies, normal forms, and examples & Text, tables, equations & 50 & 4 & 6.25/10 & 2 to 9.5 \\
6 & Database Systems, keys & Schema, instance, database keys, and relation-specific identification & Text, tables, equations & 30 & 4 & 6.37/10 & 3 to 10 \\
7 & Computer Networks, OSI model & Seven layers, technical accuracy, protocols, and presentation & Text, optional diagrams & 70 & 5 & 6.46/10 & 3 to 10 \\
8 & Data Mining, KDD & Definition, KDD steps, importance, and application & Plain text & 42 & 4 & 9.02/16 & 0 to 16 \\
9 & Algorithms & Sorting choice, negative-cycle detection, and dynamic programming & Text, equations, diagrams & 83 & 4 & 13.28/16 & 7 to 16 \\
10 & Fisheries & Climate, climate variables, and temperature-related changes in rohu & Plain text & 23 & 3 & 5.49/6 & 3.75 to 6 \\
11 & Object Oriented Programming & Method overloading, overriding, comparison, and examples & Text and code & 19 & 4 & 8.55/16 & 4 to 15 \\
12 & E-commerce & Encryption, firewall, and symmetric-key encryption & Plain text & 5 & 3 & 4.40/9 & 0 to 8 \\
\bottomrule
\end{tabularx}
\end{table*}

\begin{figure*}[!t]
\centering
\begin{subfigure}[t]{0.32\textwidth}
\centering
\includegraphics[width=\linewidth,trim=18 340 175 30,clip]{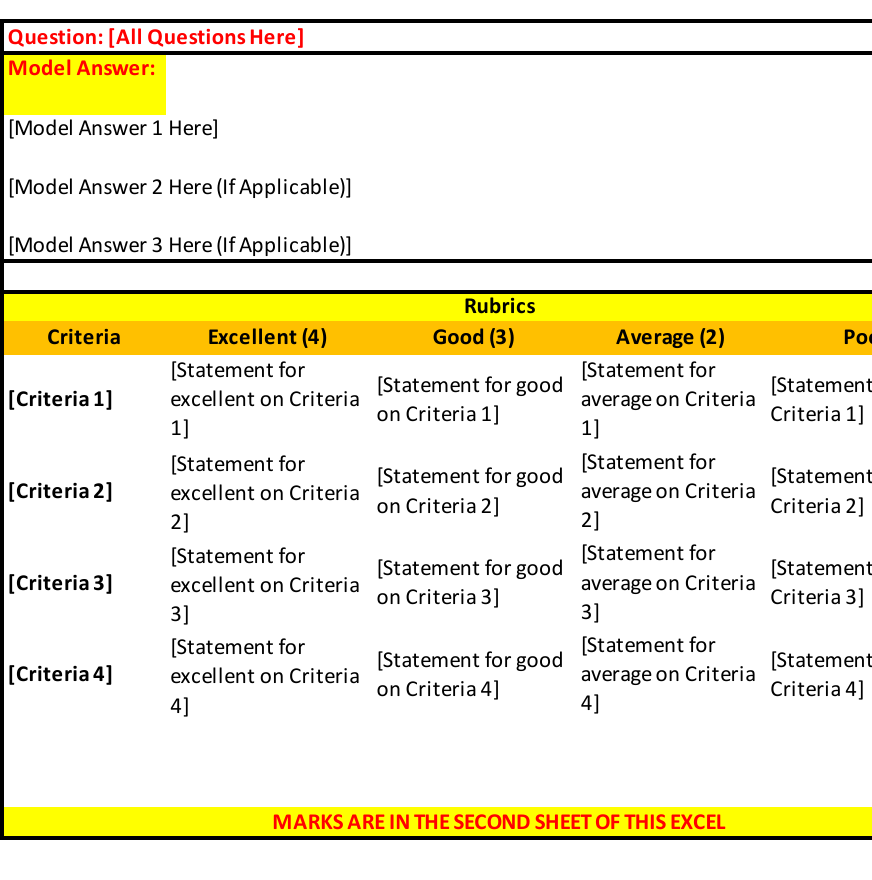}
\caption{Illustrative blank rubric template.}
\end{subfigure}\hfill
\begin{subfigure}[t]{0.32\textwidth}
\centering
\includegraphics[width=\linewidth,trim=18 90 95 25,clip]{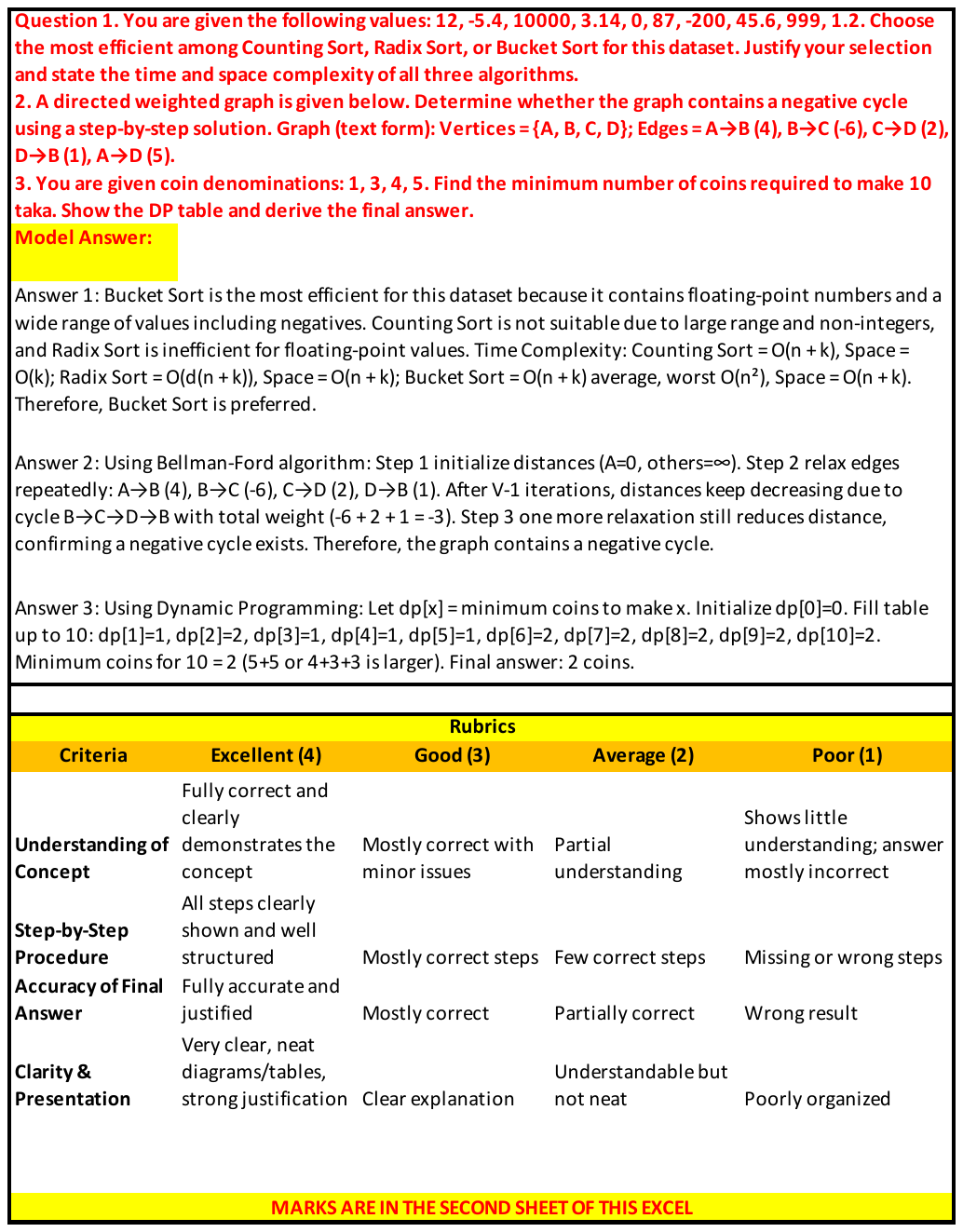}
\caption{Illustrative completed question, model answers, rubric, and criterion-level marks.}
\end{subfigure}\hfill
\begin{subfigure}[t]{0.32\textwidth}
\centering
\includegraphics[width=\linewidth,trim=18 270 140 30,clip]{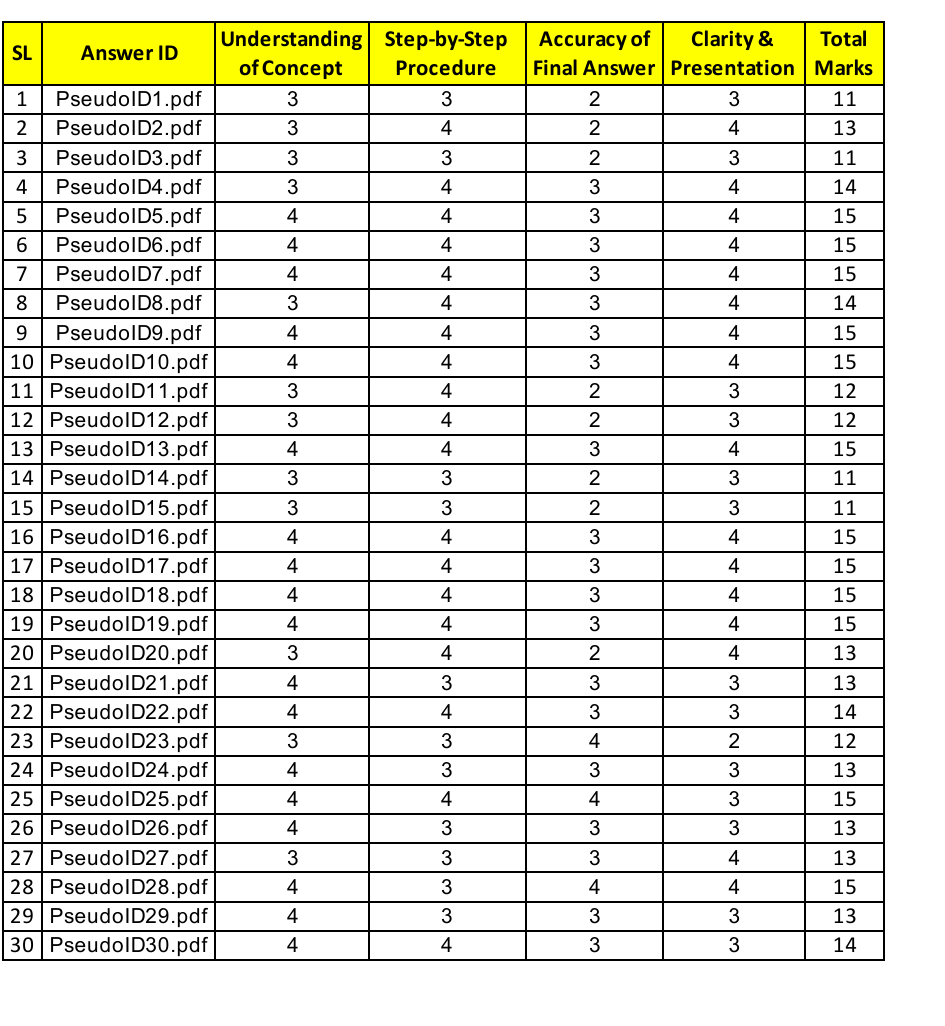}
\caption{Criterion marks and total marks.}
\end{subfigure}
\caption{Outcome-Based Education rubric materials used during data acquisition. The three source PDFs are included directly and displayed as subfigures.}
\label{fig:rubrics}
\end{figure*}

Rubrics were defined at the question level. Each criterion names an assessable component and associates it with textual descriptions of performance quality. The supplied template uses four performance columns. Numeric allocations remain question-specific, and the final mark for a criterion is stored in \texttt{gained\_marks}. Performance labels provide descriptive guidance; they are not a substitute for the criterion maximum. Figure~\ref{fig:rubrics} presents the blank template and the two pages of one completed rubric example as requested.

\section{Experimental Design, Materials and Methods}

\subsection{Data acquisition and expert rubric design}

The source data were collected through participating faculty members at four academic institutions. A total of eight faculty contributors supplied scanned examination scripts and examiner-prepared assessment spreadsheets. Student consent was obtained for preparation and research use of the scripts. The source pool represented 415 students and produced 485 answer submissions.

Faculty contributors provided the question, model or reference answer, rubric, criterion descriptions, criterion-level marks, and total mark for each applicable answer. Rubrics were designed for specific questions with involvement from an OBE expert. Criteria reflected the knowledge or performance components expected in each response, including conceptual correctness, completeness, procedural accuracy, examples, technical accuracy, interpretation, and clarity. Different questions retained different numbers of criteria and mark allocations.

The scanned PDFs were preserved rather than transcribed into text-only responses. This decision retained handwriting, page layout, equations, tables, sketches, diagrams, code, and other evidence that may be relevant to multimodal assessment. The original scripts also retained realistic differences in page count and question organization.

The answer scripts were digitized through multiple acquisition pathways. Mobile scanning applications, including CamScanner and Adobe Scan, were used alongside conventional document scanners. The resulting files retain variations in illumination, brightness, contrast, shadows, orientation, cropping, compression, and resolution. Student responses also exhibit diverse handwriting, overwritten text, crossed-out work, inserted corrections, and revised calculations. Preserving these traits broadens the visual domain of the collection and supports evaluation across realistic acquisition and writing conditions.

\subsection{Consolidation of heterogeneous examiner data}

The examination materials were collected from different faculty members and institutions, and the raw files were not initially available in one standardized representation. Differences existed in file naming, spreadsheet organization, question formatting, criterion formatting, and mark representation. The first processing operation combined these independently submitted sources into a common answer-level data structure.

For each student answer, the corresponding PDF was linked with:
\begin{itemize}
\item subject information;
\item examination question;
\item model or reference answer;
\item OBE rubric criteria;
\item rubric-level performance descriptions;
\item criterion-level obtained marks; and
\item total obtained marks.
\end{itemize}

The merged representation was organized with one record for each complete answer submission. Figure~\ref{fig:json} shows the resulting JSON pattern. This organization preserves the relationship among the visual answer, academic prompt, reference material, rubric definitions, and examiner-assigned marks.

The consolidation process established a one-to-one mapping between each metadata object and one scanned PDF. Rubric objects were stored as arrays to preserve question-specific criterion counts. No fixed four-criterion structure was imposed. This choice retained the three-criterion Fisheries and E-commerce rubrics and the five-criterion Computer Networks rubric.

\subsection{Metadata standardization and completion}

Subject labels from examiner files were mapped to the nine finalized names reported in Table~\ref{tab:subjects}. Capitalization, spacing, and common abbreviations were standardized. Question strings were cleaned to remove redundant score annotations where they did not form part of the academic prompt. Question numbers and ordered subquestions were retained. Criterion text received typographic and spacing corrections while its educational meaning and original allocation were preserved.

\subsection{Score validation}

Every criterion mark was checked against the applicable criterion maximum. Any source value above its valid maximum was corrected during preparation, and the total mark was updated where the correction changed the aggregate. The final answer-level JSON was audited across all 485 records. Each stored total equals the sum of its criterion marks. No total exceeds its question-level rubric maximum. The verified range for Question 1 is 0 to 5, with a mean of 2.82 marks.

Question maxima vary across the collection. Normalized descriptive marks in Table~\ref{tab:subjects} were calculated for each answer as
\begin{equation}
s_i = \frac{m_i}{M_{q(i)}},
\end{equation}
where $m_i$ is the total mark for answer $i$ and $M_{q(i)}$ is the maximum for its question. A subject mean is the arithmetic mean of $s_i$ over all answers in that subject. The question-level table retains the original mark units.

\subsection{De-identification and referential integrity}

The 485 records were assigned a randomly shuffled set of unique integers from 0 to 484. Original filenames were replaced by randomly generated unique PDF filenames. The \texttt{pdf\_file} value was updated at the same time as the physical filename to preserve the correct mapping.

Integrity checks verified answer-ID uniqueness, filename uniqueness, one PDF reference per record, and preservation of the pre-randomization record-to-file mapping. The final metadata contain 485 unique IDs and 485 unique randomized filenames. The file preparation process checked that each referenced PDF existed in the assembled file directory.

Record-level de-identification does not guarantee removal of text or logos embedded within every scan. Visible identifiers that remain in image content are addressed through restricted research access and the privacy conditions described in Section~\ref{sec:ethics}.

\subsection{Preparation workflow}

\begin{figure*}[htbp]
\centering
\includegraphics[width=\textwidth]{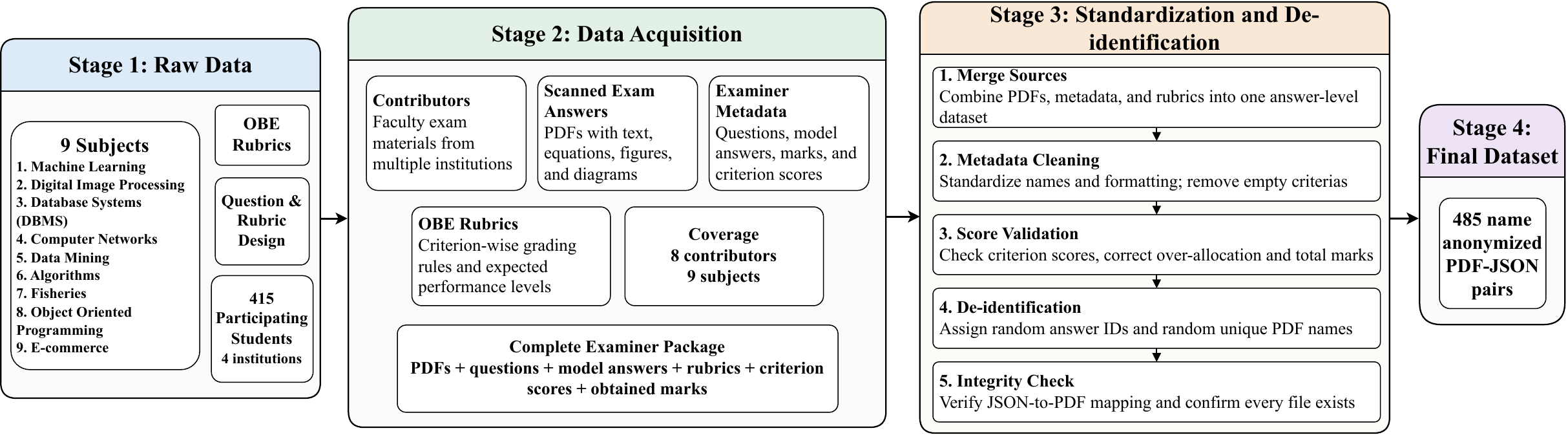}
\caption{Dataset preparation workflow from raw examination materials to the standardized answer-level collection. The supplied process diagram is included directly.}
\label{fig:workflow}
\end{figure*}

Figure~\ref{fig:workflow} summarizes the four-stage process. Raw questions and rubric designs were combined with contributions from 415 students. Data acquisition linked scans, examiner metadata, and OBE rubrics. Standardization merged sources, cleaned metadata, validated marks, randomized IDs and filenames, and checked integrity. The final organization contains 485 answer-level PDF-JSON pairs.

\subsection{Suggested research use}

A typical research pipeline can load a JSON object, retrieve its paired PDF, render one or more pages, and construct a target from the total mark or criterion marks. Models may consume page images, optical-character-recognition text, layout features, or multimodal combinations. Training and evaluation partitions should be created at the question level or with grouped controls when the research objective involves transfer to unseen questions. Random answer-level splitting can place visually related responses to the same prompt in every partition and may overstate cross-question generalization.

Criterion marks can be represented in their original units or divided by criterion maxima. Question and subject labels should remain available for stratification. The smallest subject contains five answers, while the largest contains 88. Sampling plans and uncertainty estimates should reflect this imbalance. Any generated feedback should be treated as a research output and must not be presented as an official grade.

The paired files and machine-readable metadata support principles of interoperability and reuse \citep{wilkinson2016}. The restricted access condition reflects the presence of educational records and residual identifiers in some source images.

The naturally occurring image variation also supports generalization-oriented evaluation. Robustness studies can group pages by observable characteristics such as illumination, contrast, skew, compression, handwriting style, correction density, and document complexity. The combination of mobile scanning applications and conventional scanners creates acquisition shifts that are useful for assessing whether a model remains stable outside a single capture pipeline. These variations strengthen the dataset as a research resource for evaluating multimodal document understanding under realistic examination conditions.

\section{Limitations}

The collection has several constraints relevant to reuse. It covers nine subjects and only 12 question templates, and its subject distribution is uneven. E-commerce has five answers, while Machine Learning has 88. Question-specific rubrics use different maxima, criterion counts, and textual scales. Comparisons across questions require normalization or question-aware modeling.

\section{Ethics and Privacy Statement}
\label{sec:ethics}

Student consent was obtained for preparation and research use of the examination scripts. The dataset and associated research activities were separated from official grading decisions. The data were not used to determine, alter, or replace the direct grades assigned to participating students.

Answer IDs and PDF filenames were randomized during preparation. Some scanned pages retain student names, student IDs, university names, institutional logos, or related identifiers within the page image. These elements are retained only within the controlled research context. Dataset access is limited to approved research purposes. Recipients must protect the files, avoid re-identification, avoid public redistribution, and report only aggregate or de-identified information.

Any future use involving automated assessment must retain human monitoring. Outputs from research models must not be treated as official grades or used as the sole basis for a decision affecting a student.

\section*{Acknowledgments}

This project was funded by the Bangladesh Accreditation Council (BAC) under the project \textit{AI-Powered Automated Exam Evaluation and OBE-Based University Assessment System}. The authors acknowledge the participating students, faculty contributors, institutions, and the OBE expert who supported rubric preparation and data collection.

\section*{Data and Code Availability}

The dataset is deposited on Zenodo \href{https://doi.org/10.5281/zenodo.22058761}{(DOI: 10.5281/zenodo.22058761)} under restricted access and is available for research use upon reasonable request via the repository, subject to privacy review and the safeguards stated in this article. No source code is distributed with the manuscript package; supporting preparation code may be shared by the corresponding author under the same research-only conditions.

\section*{Declaration of Generative AI and AI-Assisted Technologies in the Manuscript Preparation Process}

Generative artificial intelligence was used to assist with preparation of the manuscript. The human authors monitored and reviewed the manuscript, numerical statements, tables, figures, and references. The authors take full responsibility for the final content.

\section*{Declaration of Competing Interest}

The authors declare that there are no known competing financial interests or personal relationships that could have appeared to influence the work reported in this article.

\bibliographystyle{elsarticle-num}
\bibliography{References}

\end{document}